\documentclass[letterpaper, 10 pt, conference]{ieeeconf} 
\usepackage{microtype}
\usepackage{subfigure}
\usepackage{float}
\usepackage{booktabs} 
\usepackage{hyperref}
\usepackage{algorithm}
\usepackage{algorithmic}
\usepackage{xcolor}  

\usepackage{amsmath}
\usepackage{amssymb}
\usepackage{mathtools}
\usepackage{times}

\usepackage[capitalize,noabbrev]{cleveref}

\usepackage[textsize=tiny]{todonotes}

\usepackage[ruled,vlined,linesnumbered,algo2e]{algorithm2e}
\SetKwBlock{Finally}{finally}{end}

\usepackage{amsmath,amsfonts,bm}

\def\eqref#1{equation~\ref{#1}}

\def\1{\bm{1}}

\DeclareMathAlphabet{\mathsfit}{\encodingdefault}{\sfdefault}{m}{sl}
\SetMathAlphabet{\mathsfit}{bold}{\encodingdefault}{\sfdefault}{bx}{n}

\usepackage{hyperref}
\usepackage{url}
\usepackage{graphicx}
\usepackage{placeins}
\usepackage{amsmath}
\usepackage{cuted}
\usepackage{mathrsfs}
\usepackage{placeins}

\usepackage{booktabs}
\usepackage{longtable}
\usepackage{siunitx}
\usepackage{multirow}
\usepackage{caption}
\usepackage{amssymb}

\newcommand{\psnshort}{\textsc{PSN}}
\newcommand{\psn}{\textsc{Proximal State Nudging}}

\IEEEoverridecommandlockouts                              

\title{\LARGE \bf
Proximal State Nudging: Reducing Skill Atrophy from AI Assistance
}

\author{
Megha Srivastava$^{1}$,
Jonathan Ouyang$^{2}$,
Eric Zhou$^{2}$,
Andrew Silva$^{3}$,
Emily Sumner$^{3}$, \\
Dorsa Sadigh$^{1}$,
Yuchen Cui$^{2}$,
Deepak Gopinath$^{3}$,
and Guy Rosman$^{3}$%
\thanks{$^{1}$Stanford University, Stanford, CA, USA.
        {\tt\small meghas@stanford.edu}}%
        \thanks{$^{2}$University of California Los Angeles, Los Angeles, CA, USA.}%
\thanks{$^{3}$Toyota Research Institute, Los Altos, CA, USA. Any opinions, findings, and conclusions expressed in this material are those of the authors and do not necessarily reflect the views of TRI or any other Toyota entity.}%
}

\begin{document}

\maketitle
\thispagestyle{empty}
\pagestyle{empty}
\begin{strip}
\centering
\includegraphics[width=0.88\linewidth]{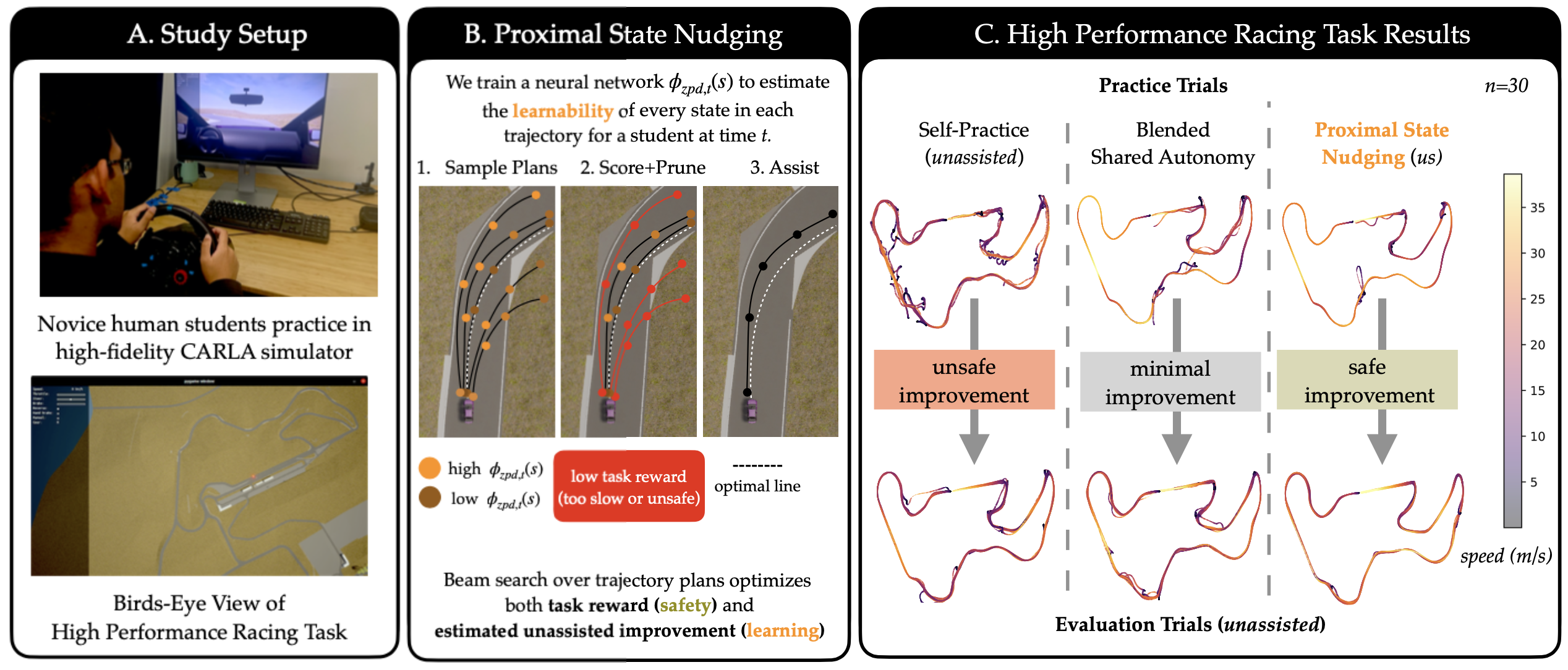}
\captionof{figure}{\textbf{Overview.} We propose \psn{} (\psnshort{}),  a learning-aware shared autonomy algorithm that minimizes human skill atrophy in rapid-control domains, such as High Performance Racing (A). Inspired by the Zone of Proximal Development theory from cognitive psychology \cite{vygotsky1978mind}, \psnshort{} maintains an estimator $\phi_{zpd,t}(s)$ of each state's  ``learnability'' -- how likely it is to improve a student's unassisted skills -- which feeds a beam search-based planner that optimizes for both learning and task reward (B). When used during student practice, \psnshort{} eliminates the tradeoff between safe performance and learning; it ensures safer trajectories with fewer collisions in comparision to self-practice without any AI assistance, while resulting in a larger improvement in task reward than standard blending-based shared autonomy (C). }
\label{fig:pull}
\end{strip}

\begin{abstract}
\emph{Skill atrophy}, the gradual decline of human capability under AI assistance, poses a safety risk in shared-control of semi-autonomous systems, where operators may be unable to distinguish their own inputs from autonomous corrections. We propose Proximal State Nudging (\psnshort{}), a shared autonomy algorithm that jointly optimizes for skill development and task performance by nudging users toward states estimated to be most learnable. 
We first show that \psnshort{} outperforms existing shared autonomy baselines in balancing student improvement in unassisted reward with overall shared performance, using simulated students in the classic LunarLander environment.   We then present, to the best of our knowledge, the first human subject studies of a planner incorporating learning-compatible shared autonomy: across two driving tasks in the CARLA simulator (High Performance Racing and Parallel Parking, n = 60), \psnshort{} produces up to 7x larger gains in unassisted skill than standard blended shared autonomy, while incurring $~50\%$ fewer collisions than unassisted self-practice.

\end{abstract}

\section{Introduction}
Skill atrophy resulting from AI assistance  has been documented across a wide range of domains, including surgery~\cite{sarofim2024devil} and transportation~\cite{casner2016challenges, de2023shared}. Despite this growing evidence, relatively little work has focused on designing AI assistance that actively supports continued human skill improvement. Instead, existing discourse frames the problem as a binary choice: either rely on AI for strong task performance, or completely remove AI support to encourage learning~\cite{dell2023chatgpt}. This framing raises the  question: \textit{Can we shape AI assistance to support task performance and safety in the moment, while also improving user's long-term skill development?}

Imagine a student pilot learning how to control an aircraft, or a novice operator learning to teleoperate a robotic manipulator. Human control of such tasks matters for two reasons: first, in high-skill recreational and professional domains -- performance driving, surgical training, and rehabilitation — human skill is itself the goal, and AI assistance serves to scaffold its acquisition. Second, human operators might still intervene when automation fails in situations like aircraft control or driving ~\cite{endsley2017autonomous}. In both cases, while AI assistance can improve task performance and safety, we argue that shared autonomy systems should also account for how assistance affects users' long-term skill retention. Few prior works have explored balancing task performance with skill maintenance, but these approaches often make task-specific assumptions (e.g. access to near-optimal Q-values), and are not grounded in principles from cognitive psychology that provide generalizable theories of how assistance impacts human learning and retention~\cite{wada2019simultaneous,bragg2020fake}.

We propose \psn{} (\psnshort{}), a learning-aware assistance framework that explicitly accounts for its influence on human skill development. \psnshort{} is inspired by the Zone of Proximal Development (ZPD~\cite{vygotsky1978mind}) from educational psychology and aims to estimate the ``learnability'' of states—how conducive they are to supporting user learning—across the environment. For example, in a high performance racing task, a student driver on the racing line might carry a turn cleanly but struggle to recover after drifting wide, making off-line states the most fruitful for skill growth (Figure~\ref{fig:pull}). \psnshort{}  optimizes over future actions to balance coverage of such high-learnability states with standard shared autonomy objectives, including safety and task performance.

Crucially, \psnshort{} estimates learnability without requiring an explicit model of how assistance changes human learning or task-specific assumptions such as access to near-optimal Q-values. Instead, it draws on the ZPD principle that learning is most productive when users face challenges matched to the boundary of their current ability, and operationalizes this by comparing a user's estimated expected reward from a given state with and without assistance. This grounding in cognitive scaffolding theory, combined with compatibility with any underlying shared autonomy policy, allows us to deploy and evaluate \psnshort{} with real human students.  Our overall contributions include: 
\begin{enumerate}
    \item An algorithm for learning-compatible shared autonomy grounded in cognitive scaffolding
    \item Empirical validation showing that \psnshort{} improves upon known shared autonomy baselines in the classic \texttt{LunarLander} environment with simulated students 
\item Two distinct human-subject studies involving real student learners ($n=60$) in realistic driving tasks (Parallel Parking and High Performance Racing) using the open-source CARLA autonomous driving simulator \cite{dosovitskiy2017carla}, showing that \psnshort{} results in up to 7x gains in improving unassisted skill performance than standard blended shared autonomy, while also incurring $50\%$ fewer collisions than unassisted self-practice. 
\end{enumerate}

We show that by guiding users toward states that foster learning while maintaining strong task performance, \psn{} can help prevent skill atrophy.
\section{Related Works}
\psn{} lies at the intersection of shared autonomy for human-robot interaction and research on skill atrophy under automation. Shared autonomy systems combine human and AI control to support complex task execution while preserving (partial) human agency, typically assuming users are already skilled. \cite{dragan2013policy} introduced policy-blending formalisms for shared control, \cite{javdani2015shared} proposed hindsight optimization under goal uncertainty, and \cite{reddy2018shared} relaxed assumptions on goal spaces via model-free deep RL. Related ``minimum intervention'' frameworks~\cite{broad2019highly} similarly preserve human agency but do not reason about learning. 

A smaller line of work explicitly targets skill preservation: \cite{bragg2020fake} introduced Stochastic Q-Bumpers, which, unlike our work, assumes Q-learning updates for both student and expert policies and then modulates assistance based on Q-value gaps to retain learning opportunities. \cite{Fitzsimons19} and \cite{srivastava2022motor} modulate assistance for motor skill training; \cite{Agarwal19} used shared autonomy to induce variability in rehabilitation robotics, implicitly targeting high-ZPD states; and \cite{tian2023humanlearning} modeled human learning dynamics in interaction. Recently \cite{srivastava2025shared} applied assistive coaching to high-performance racing also inspired by cognitive scaffolding, but did not incorporate learning objectives explicitly into a shared autonomy planner.

Meanwhile, skill atrophy from automation over-reliance has been documented across safety-critical control domains, including aviation~\cite{casner2014impact}, driving~\cite{de-Winter2023-cp}, and surgery~\cite{sarofim2024devil}, with medical evidence that even expert performance degrades once assistance is removed~\cite{parchmann2024ai}.  Our work directly targets this gap, aligning with broader human-centered AI efforts emphasizing user empowerment~\cite{ellis2025training} while focusing specifically on sustained skill development in cyber-physical shared control.
\section{Formalism}
\label{sec:formalism}
We consider a student with an unassisted policy $\pi_{\mathrm{student},t} \in \Pi$ at each \emph{learning timestep} $t$, mapping states to actions ($s \in S \rightarrow a \in A$). Our objective is to design a assistive policy for $\pi_{\mathrm{student},t}$ that satisfies two properties:

\begin{enumerate}[noitemsep, topsep=0pt]
    \item \textbf{Safety:} Safe and effective task completion across time, measured by high reward and task performance averaged across episodes when the student is \textit{assisted}. 
    \item \textbf{Learning:} $\pi_{\mathrm{student},t}$ improves over time, measured by high reward and task performance averaged across episodes when the student is \textit{not assisted}. 
\end{enumerate}

As described earlier, most approaches to shared autonomy often fail our second goal, as providing useful assistance might result in the student incorrectly attributing high reward to their own incorrect input. However, determining how to structure assistance that actually promotes student learning remains challenging due to the lack of a strong model of human adaptation, and prior works on modeling human learning make strong assumptions such as using Wiener processes or linear-quadratic dynamics for the student model \cite{yu2023coach, tian2023humanlearning}.

In contrast, \textsc{Proximal State Nudging} adopts a state-centric view of skill learning, inspired by Vygotsky’s Zone of Proximal Development (ZPD) to adapt instruction difficulty \cite{vygotsky1978mind}. We estimate $\phi_{\mathrm{zpd}, t}$, the likehood that a state $s \in S$ supports the student's learning at a given point in time $t$. States that are too difficult (low chance of high reward) or too easy (already mastered) should have low learnability. The shared policy  nudges the student toward states with higher $\phi_{\mathrm{zpd}, t}$ by selecting actions that balance task optimality and state learnability. We now expand on \psnshort{}.  

\subsection{Shared Control Planning for Assistance}
\psn{} can build upon any existing shared control strategy $\pi_\text{shared}(a|s)$ which, given the student policy $\pi_\text{student,t}(a|s)$ and an expert agent policy $\pi_\text{expert,t}(a|s)$, selects the next action for a given level of assistance $\alpha$. For example, a standard form of assistance is blending-based shared autonomy, which samples from a convex combination of the optimal agent and student policies, controlled by $\alpha$ (\cite{wang2020review, doi:10.1177/154193120104502323}):
\begin{equation}
\label{eq:stdsa}
\begin{aligned}
\pi_{\mathrm{shared}}(a \mid s)
&\triangleq
\alpha\,\pi_{\mathrm{expert}}(a \mid s) \\
&\quad
+ (1-\alpha)\,\pi_{\mathrm{student}}(a \mid s),
\qquad \alpha \in [0,1].
\end{aligned}
\end{equation}

During planning, \psnshort{} modifies
$\pi_{\mathrm{shared}}$ by first employing a beam search over a set of
length-$T$ sampled trajectories
$\mathcal{B}_t(\hat{s}) \sim \pi_{\mathrm{shared}}(\cdot \mid \hat{s})$,
starting at the current student's state $\hat{s}$. We then select the first
action of the sampled trajectory that maximizes the score $J$, defined as a
weighted sum of average predicted state learnability and task reward:
\begin{equation}
\begin{aligned}
J_t(\tau)
&\triangleq
w_1\!\left(\frac{1}{T}\sum_{i=0}^{T-1}\phi_{\mathrm{zpd}, t}(s_i)\right)
+
w_2\,R(\tau), \\
a_{\mathrm{PSN},t}(s)
&\triangleq
a_0(\tau^\star),
\qquad
\tau^\star \in \arg\max_{\tau \in \mathcal{B}_t(s)} J_t(\tau).
\end{aligned}
\label{eq:psn}
\end{equation}

Here, $R$ is the task reward function, and $\phi_{\mathrm{zpd}, t}(s_i)$ is the  estimated learnability of state $s_i$, which we discuss next.

\subsection{Proximal State Estimation: Identifying Learnable States}
The key mechanism of \textsc{Proximal State Nudging} is estimating a state \textit{learnability} function $\phi_{\mathrm{zpd},t}(s)$, which operationalizes the concept of ZPD, or the gap between what tasks a student can do independently and with assistance,  from cognitive psychology  \cite{vygotsky1978mind}. Given a current student's state $\hat{s}$, our goal is to estimate whether there is room for improvement in task performance given the student's current skill level, which is naturally upper-bounded by the performance achievable under expert assistance. Accordingly, if we sample trajectories starting from the student's state $\hat{s}$ with ($\pi_{\mathrm{shared}}$) and without ($\pi_{\mathrm{student},t}$) assistance, we  can approximate $\phi_{\mathrm{zpd}, t}(\hat{s})$ as the difference in expected task reward:
\begin{equation*}
  \mathbb{E}_{\tau \sim \pi_{\mathrm{shared}}, s_0 =\hat{s}}[R_{\mathrm{task}}(\tau)] - \mathbb{E}_{\tau \sim \pi_{\mathrm{student},t}, s_0 =\hat{s}}[R_{\mathrm{task}}(\tau)].
\end{equation*}

For efficiency, we reuse the student's training rollouts from $s_\textit{start}$ to train $\phi_{\mathrm{zpd},t}$, treating each visited state along a trajectory as a candidate $\hat{s}$. Any state $\hat{s}$ from which the expected task reward is similar with or without assistance --- either because they are too easy or too hard for the student agent --- will receive low $\phi_{\mathrm{zpd}, t}(\hat{s})$.  Furthermore, the $\phi_{\mathrm{zpd}, t}$ estimator can be updated at each learning step $t$ (or at a fixed interval), thus evolving with the student policy.  In practice, $\phi_{\mathrm{zpd}, t}(\hat{s})$ is normalized to be between 0 and 1 via the sigmoid function, and function approximation can be used for large  state dimensions. By not requiring any explicit student model, \psn{} can be used in  settings with real human student learners, unlike prior work (Section \ref{sec:human}). 

\subsection{Algorithm Overview and Adaptive $\alpha$}
Algorithm~\ref{alg:alg1} summarizes our full approach for \psn{}. \textsf{UpdateZPD} trains the $\phi_{zpd,t}(s)$ estimator at time $t$ on states where the student is either assisted and unassisted for the entire episode; its sample complexity can be reduced by using a small model size. While in simulation \textsf{UpdatePolicy} can be implemented using any standard reinforcement learning algorithm, our primarary goal is to support human learners, where this function is implicit and likely task-dependent.  Finally, while we write Algorithm~\ref{alg:alg1}  as if the \textsf{BeamSearch} procedure happens at every state during training, in practice this can be implemented at fixed interval to reduce latency. 

The hyperparameter $\alpha$ typically denotes a fixed level of assistance in shared autonomy methods \cite{reddy2018shared}. However,
students can only attribute observed outcomes to their own actions when assistance is low, so we taper $\alpha$ in a way that is proportional to the estimate $\phi_{zpd,t}(s)$ of the current state. This encourages student learning in the high learnable states that $\psn{}$ pushes the student towards. 

\begin{algorithm}
\caption{Proximal State Nudging}
\label{alg:proximal_state_nudging}
\small
\begin{algorithmic}[1]
\STATE \textbf{Input:} Task environment with state space $\mathcal{S}$, action space $\mathcal{A}$
\STATE \hspace{1.5em} Expert policy $\pi_{\mathrm{expert}}$
\STATE \hspace{1.5em} Base shared policy $\pi_{\mathrm{shared}}$
\STATE \hspace{1.5em} Assistance level $\alpha \in [0,1]$
\STATE \hspace{1.5em} Scoring weights $w_1, w_2 \in [0,1]$
\STATE \hspace{1.5em} Planning horizon $T$
\STATE \hspace{1.5em} Beam search size $B$
\STATE \hspace{1.5em} Task reward function $R_{\mathrm{task}}$
\STATE \textbf{Initialize:} Student policy $\pi_{\mathrm{student},0}$, empty datasets $\mathcal{D}_{\mathrm{shared}}$ and $\mathcal{D}_{\mathrm{student}}$ , proximal state estimator $\phi_{\mathrm{zpd}, t}$
\FOR{learning timestep $t$}
    \STATE \textcolor{olive}{\textit{// Shared control execution}}
    \FOR{each encountered state $s$}
        \STATE $\alpha' \leftarrow \alpha * (1- \phi_{\mathrm{zpd}, t}(s))$
        \STATE $\mathcal{B}_t(s) \leftarrow \textsf{BeamSearch}(\pi_{\mathrm{shared}(a|s, \alpha')}, s, T, B)$
        \FOR{each trajectory $\tau = (s_0, a_0, \ldots, s_{T-1}) \in \mathcal{B}_t(s)$}
            \STATE $J_t(\tau) \leftarrow w_1\!\left(\frac{1}{T}\sum_{i=0}^{T-1}\phi_{\mathrm{zpd}, t}(s_i)\right) + w_2 R_{\mathrm{task}}(\tau)$
        \ENDFOR
        \STATE $\tau^\star \leftarrow \arg\max_{\tau \in \mathcal{B}_t(s)} J_t(\tau)$
        \STATE $a_{\mathrm{PSN},t}(s) \leftarrow a_0(\tau^\star)$
        \STATE Add $(s, a_{\mathrm{PSN},t}(s), r, s')$ to student experience
    \ENDFOR
    \STATE \textcolor{olive}{\textit{// Update proximal state estimator}}
    \STATE Collect assisted episodes: $\mathcal{D}_{\mathrm{shared}} \sim \pi_{\mathrm{shared}}$
    \STATE Collect unassisted episodes: $\mathcal{D}_{\mathrm{student}} \sim \pi_{\mathrm{student},t}$
    \STATE $\phi_{\mathrm{zpd}, t} \leftarrow \textsf{UpdateZPD}(\mathcal{D}_{\mathrm{shared}}, \mathcal{D}_{\mathrm{student}})$
    \STATE \textcolor{olive}{\textit{// Update student policy}}
    \STATE $\pi_{\mathrm{student},t+1} \leftarrow \textsf{UpdatePolicy}(\pi_{\mathrm{student},t}, \text{experience})$
\ENDFOR
\end{algorithmic}
\label{alg:alg1}
\end{algorithm}
\begin{figure*}
    \centering
    \includegraphics[width=0.8\linewidth]{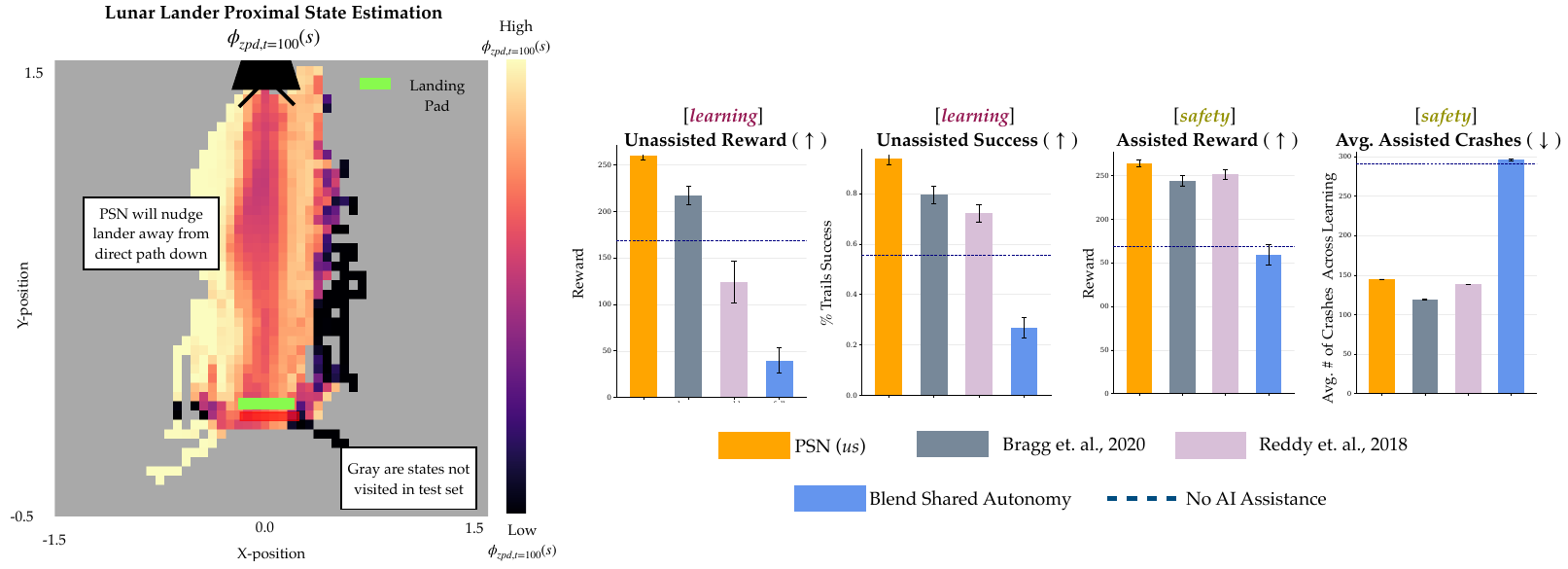}
    \caption{\textbf{Lunar Lander Task and Results}  (Left) Heatmap showing estimated proximality scores for states in \texttt{Lunar Lander}, estimated over states drawn from a test set after training for 100 episodes. States directly below the lander (too easy) or too far to the right (too challenging) are estimated to be less ``learnable''; a student agent will face similar expected reward regardless of receiving or not receiving assistance. \psnshort{} will nudge the student to states on the left, while maintaining overall task performance. (Right) \psnshort{} outperforms all shared autonomy baselines and the no-assistance baseline in learning (high unassisted reward and \% successful landings) while maintaining safety (high assisted reward and low total number of crashes during learning). Error bars show standard error across 10 random seeds and 100 episodes.}
    \label{fig:results-lander}
\end{figure*}

\section{Empirical Results: Simulated Learners}
\label{sec:sim}
We first compare \textsc{Proximal State Nudging} against existing shared autonomy baselines, using simulated students for fair comparison. Specifically, we focus on the  \texttt{Lunar Lander} (continuous state) task studied in prior work, including \cite{bragg2020fake} and \cite{reddy2018shared}. To the best of our knowledge,  \cite{bragg2020fake} is the only prior work that also seeks to improve a student's unassisted performance (\cite{reddy2018shared} and other shared autonomy works focus on improving assisted task performance). However, their proposed method, called Stochastic Q-Bumpers, assumed Q-learning style agents, which might not be applicable in real-world human learning tasks like the ones we study in Section \ref{subsec:HPR}. Nevertheless, as this is our only known baseline we follow \cite{bragg2020fake} and train Double-DQN student agents \cite{hasselt2016double}, and then analyze the pareto frontier between  assisted and unassisted task performance. These  two metrics correspond to our goal of balance safety and learning: optimizing for strong assisted task performance ensures safety of the student agent, while strong unassisted task performance demonstrates that such assistance does not result in skill atrophy.



\subsubsection{Task Overview} \texttt{LunarLander}~\cite{brockman2016openai} is a classic rapid-control task in which an agent must safely land a spacecraft in a 2D environment. The environment has a continuous 8-dimensional state space encoding position, velocity, orientation, angular velocity, and ground contact indicators, and a discrete action space of four engine commands (left, right, main, none). The multi-dimensional reward function is shaped to encourage proximity to the landing pad, low velocities, upright orientation, and successful leg contact, while penalizing fuel usage and crashes. Total rewards of $>200$ indicate strong task performance. 

\paragraph{Methods} We compare five methods:
\begin{itemize}
    \item \textbf{Bragg et. al., 2020 (Baseline)}: We implement the learning-compatible decision support algorithm from \cite{bragg2020fake}, which overrides the student's action $a_{\text{student}}$ with the expert action $a_{\text{expert}} = \arg\max_a Q^*(s,a)$ whenever the Q-value gap $\Delta Q(s) = Q^*(s, a_{\text{expert}}) - Q^*(s, a_{\text{student}})$ exceeds a threshold, with override probability proportional to $\Delta Q(s)$. 
    \item \textbf{Reddy et. al., 2018 (Baseline)}: We implement the shared autonomy algorithm introduced in \cite{reddy2018shared}, which uses model-free deep RL to train an assistive policy $\pi_{\text{shared}}$ that maximizes task reward while constraining its output to remain close to the student's input.
    \item \textbf{Blended Shared Autonomy (Baseline)}: We implement the standard policy-blending formalism from Equation \ref{eq:stdsa}. This baseline is also used in prior work  to optimize for assisted task performance, but does not explicitly account for student learning \cite{dragan2013policy}.

    \item \textbf{No AI Assistance (Baseline)}: $\pi_{student}$ is updated from experience without any form of assistance at any point.

\item \textbf{Proximal State Nudging (PSN)}: We collect unassisted and assisted rollouts from $\pi_{student}$ every 30 episodes, which are  used for both evaluation and updating $\phi_{\text{zpd},t}$. Beam search over $B=3, T=3$  trajectories is sampled from the combined $\pi_{student}$ and   $\pi_{expert}$ Q-values. We set $R$ to the task reward, and $w_1=0.5$ and $w_2=0.5$.
\end{itemize}

\subsubsection{Technical Implementation}
\paragraph{Assistive Control Design}
For all methods, we use Double DQN to train a 3-layer MLP with hidden-layer size 128 for both $\pi_{expert}$ and  $\pi_{student}$ \cite{hasselt2016double}. We train $\pi_{expert}$  to convergence for 1000 epochs at a learning rate of 0.001. We report results for assistance level $\alpha=0.1$, as we observe that higher values do not increase assistive performance, and simply deteriorate the unassisted reward of $\pi_{student}$.

\paragraph{Proximal State Estimation} We use a small 3-layer MLP, with hidden layer size 16, for $\phi_{zpd, t}$.  We update $\phi_{zpd, t}$ every 30 episodes, how which states are ``learnable'' for the student agent will change over time.To train $\phi_{zpd, t}$, we first train two separate MLPs ($\rho_{\text{sa}}$ and $\rho_{\text{ua}}$), corresponding each to the sampled assisted and unassisted episodes for evaluation, and then define the overall estimator using an affine normalization of the output difference ($\rho_{\text{sa}}$ - $\rho_{\text{ua}}$). This two-model approach is necessary due to the continuous state space result in trajectories that rarely overlap exactly. 

Figure \ref{fig:results-lander} (left) shows a heatmap of the outputs of $\phi_{zpd, t=3}$ across different states in evaluation rollouts in the \texttt{Lunar Environment}. At $t=3$, or 90 episodes, we observe that $\phi_{zpd, t=3}$ is low in the region directly below the lander's initial state, suggest that the region is ``too easy'' for the student. Likewise, states at the fringe edges (e.g. far right)  have extremely  low $\phi_{zpd, t=3}$ estimates, corresponding to challenging regions. At this point in time \psnshort{} will likely nudge to student to the middle regions that are off-center; challenging enough to improve learning, yet safe enough to recover from during learning.  
\subsubsection{Empirical Results} We compare all three methods at the end of training for 300 episodes, which is sufficient time for a student policy without assistance to start to observe successful landings. Figure \ref{fig:results-lander} shows that overall unassisted task reward and number of succesful landings  is significantly higher with \psnshort{} than all baselines, including the no AI assistance method (navy dashed line) which learns slowly. At the same time, \psnshort{} also mantains high assisted reward that is comparable to the Bragg et. al., 2020 and Reddy et. al., 2018 shared autonomy baselines. As expected, the no AI assistance baseline results in a significant number of crashes during learning, which \psnshort{} successfully mitigates. 

In addition to ensuring a successful balance of both learning and safety for the student agent, \psnshort{} holds another advantage over shared autonomy baselines: it does not assume Q-learning style updates from either the student or the expert, nor requires access to the student internal policy (e.g.Q-values). \textit{What happens when we use \psnshort{} with real human students, who may not follow a straightforward learning strategy?}

\section{Empirical Results: Human Subjects}
\label{sec:human}

\begin{figure*}
    \centering
    \includegraphics[width=0.8\linewidth]{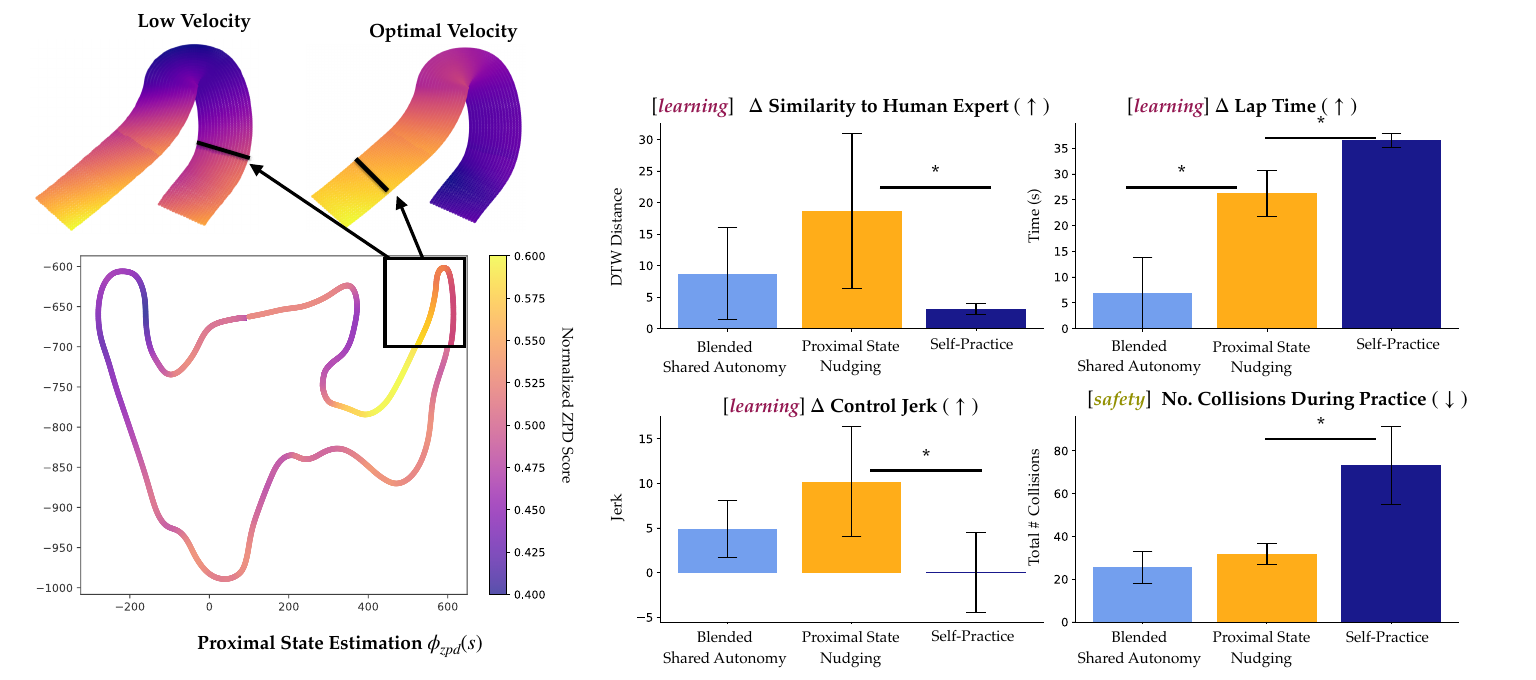}
    \caption{
    \textbf{High Performance Racing Results} (Left) Heatmap  $\phi_{\text{zpd, t=1}}$ estimates shows that assistance is predicted to most strongly support student learning at states near high-speed tight turns, and that predicted learnability is sensitive to both location and velocity. (Right) \psn{}  balances the trade-off between learning gains (e.g. improved similarity to expert, lap time, and control jerk) and safety during practice, in comparison to both blended shared autonomy and unassisted self-practice baselines.  Bars report mean standard error,  stat. sig. determined with  Welch’s t-test.}
    \label{fig:results_racing}
\end{figure*}

We next evaluate how \textsc{Proximal State Nudging} supports real human users learning two dynamical control tasks: Parallel Parking and High Performance Racing. We built environments for both tasks using the CARLA simulator~\cite{dosovitskiy2017carla}, often used for autonomous driving research~\cite{li2023learning, srivastava2025shared, costa2025coach, mihaylova2025hri}. We ask two research questions: 

\begin{enumerate}
    
    \item \textbf{RQ1 (Learning)} Does \textsc{Proximal State Nudging} improve learning (change in unassisted task performance)? We hypothesize that PSN will outperform \textit{blended shared autonomy}, and comparable to a \textit{self-practice} baseline. 
     \item \textbf{RQ2 (Safety)} Does \psn{} maintain strong overall task performance during learning? We hypothesize that \psnshort{} will outperform \textit{self-practice}, while stay comparable to \textit{blended shared autonomy}.  
\end{enumerate}

Our unifying hypothesis is that \textsc{PSN} strikes a balance between learning gains (unassisted) and safe task performance during learning (assisted) when compared to baselines. Because both tasks are sufficiently complex that learning expert policies with RL is challenging, the only baseline from Section \ref{sec:sim} that we can compare \psnshort{} against in this section is the Blended Shared Autonomy baseline. Furthermore, as we anticipate that learning is a slow temporal process for human students,  we  build our $\phi_{zpd}$ estimator using offline data, assuming we can treat state learnability as fixed for the 1 hour duration of our experiments.  We recruit $n=60$ participants, mostly university students, across two institutions. Our study was approved by our Institutional Review Board.

\subsection{High Performance Racing}
\label{subsec:HPR}
In the high-performance racing (HPR) task, users are instructed to drive around a fixed race track with the goal of minimizing their overall lap time.
This task is particularly well-suited for studying human skill acquisition because it has clear success metrics for both learning (e.g. improved lap time) and safety (e.g. total number of collisions).

\subsubsection{Task Overview}
We follow the HPR task set-up studied in several prior works \cite{decastro2024dreamingassistlearningalign, lidard2024blending, srivastava2025shared, sumner2025simcoachcorpusnaturalisticdatasetlanguage}. Specifically, we simulate the 2-mile track in Thunderhill Raceway Park, California in the CARLA simulator, using a vehicle configured for high performance racing. Each state in the HPR task consists of the vehicle's position ($x$, $y$, $z$), angle ($yaw$, $pitch$, $roll$), and speed ($vx$, $vy$, $vz$). The continuous action space consists of steering angle, throttle, and brake control. Each episode ( a single lap around the raceway) is evaluated with three metrics: lap time, Dynamic Time Warp distance from an expert demonstration, and smoothness (measured as jerk action input).

\paragraph{Methods}
   We compare three methods: 
\begin{itemize}
    \item \textbf{Blended Shared Autonomy (Baseline)}: We  blend the inputs of $\pi_{student}$, the user, and $\pi_{expert}$,  built from  an expert human demonstration, with $\alpha=0.8$ following Equation  \ref{eq:stdsa}. 
    \item \textbf{Proximal State Nudging (\psnshort)}: We sample, at fixed intervals along the track, the nearest point for each of $B=5$ distinct human expert trajectories, which cover different near-optimal task strategies. We then set the point in each trajectory that crosses the start of the next interval as the goal for CARLA's Global Route Planner (GRP), which returns a set of waypoint trajectories $\mathcal{B}_t(s)$. Following Equation \ref{eq:psn},  we set $R$ to the inverse trajectory length (encouraging speed),  and  $w_1=0.5$ and $w_2=0.5$. 
    \item \textbf{Unassisted Self-Practice (Baseline)}: The student repeatedly practices high performance racing around the track with full control of the vehicle. \end{itemize}

\subsubsection{Technical Implementation}

\paragraph{Assistive Control Design}
The assistive control for both Blended Shared Autonomy and \psnshort{} use human expert demonstrations provided by \cite{srivastava2025shared}. Given a point in an expert trajectory, the Path Planner in the CARLA simulator generates waypoints to input into a PID controller with the driver's state, and the control outputs are used as the actions for $\pi_{expert}$.

\paragraph{Proximal State Estimation}
 To avoid introducing latency, we used a Linear Regression model for $\phi_{zpd}$. This is suitable as  the state space is low-dimensional. To train $\phi_{zpd}$, we collect an offline dataset of 50  demonstrations of novice drivers both with and without assistance (at $\alpha=0.8$). We follow the same two-model approach, based on assisted and unassisted trajectories, described in Section \ref{sec:sim}.

Figure \ref{fig:results_racing} (left) shows a heatmap of the outputs of $\phi_{zpd}$ across different states across the Thunderhill Raceway track, showing that the states with the highest estimated learning potential for students are concentrated at the tight turn towards the end of the track. Furthermore, $\phi_{zpd}(s)$ is very sensitive to the velocity values in $s$, suggesting that the student's speed affects its learnability. Interestingly, $\phi_{zpd}(s)$ changes laterally across the track, and \psnshort{} would nudge a slower student to take the tight turn widely, while a faster student would be encouraged to focus on states after the turn's apex.  

\subsubsection{Human Subject Study Results}
We compare all three methods with the same sequence of trials:
\begin{enumerate}[noitemsep, topsep=0pt]
    \item 2 \textbf{baseline trials} with unassisted user control.
    \item 2 \textbf{practice trials} with either (i)  Blended Shared Autonomy,  (ii) Unassisted Self-Practice or (iii) \psnshort{}.
    \item 2 \textbf{evaluation trials} with unassisted user control.
\end{enumerate}

In each trial, subjects were instructed to complete one lap as quickly as possible (under 3 minutes). We recruited 30 students, evenly split between methods. We compute the three learning metrics (lap time,  Dynamic Time Wrap distance to a human expert, and control input jerk) as the difference between average baseline and evaluation performance. For safety, we measure the total number of collisions during practice trials, with lower values indicating safer learning.

Our results, reported in 
Figure~\ref{fig:results_racing}, show that \psn{} outperforms Blended Shared Autonomy across all three learning metrics, with a statistically significant difference in lap time ($\sim$ 5x), supporting our hypothesis for \textbf{RQ1}. \psnshort{} was also significantly stronger than  Unassisted Self-Practice for both similarity to expert and jerk metrics, though self-practice led to a significantly stronger improvement in lap time. Finally, users assigned either \psnshort{} or Blended Shared Autonomy had low number of collisions during practice, and significantly lower than Unassisted Self-Practice  ($<$ 50\%), validating our hypothesis for \textbf{RQ2}.

One reason why self-practice leads to a significant improvement in lap time, but not other metrics, might be that students are fixated on improving the lap time and  ignore other aspects of good driving, like smooth control. In fact, participants assigned \psnshort{}   during practice noted learning specific turning skills (e.g., \textit{``I learned there are certain chokepoints where you need to reduce speed to avoid spinning out''} and \textit{``learning when to ease off the gas pedal when heading into turns''}),.  Furthermore, self-practice students experienced a high number of collisions during practice.  These results support our hypotheses that \psnshort{}  can improve student learning (\textbf{RQ1}) while still providing enough assistance to ensure safety (\textbf{RQ2}).

\subsection{Parallel Parking}

\begin{figure*}[h]
    \centering
    \includegraphics[width=0.8\linewidth]{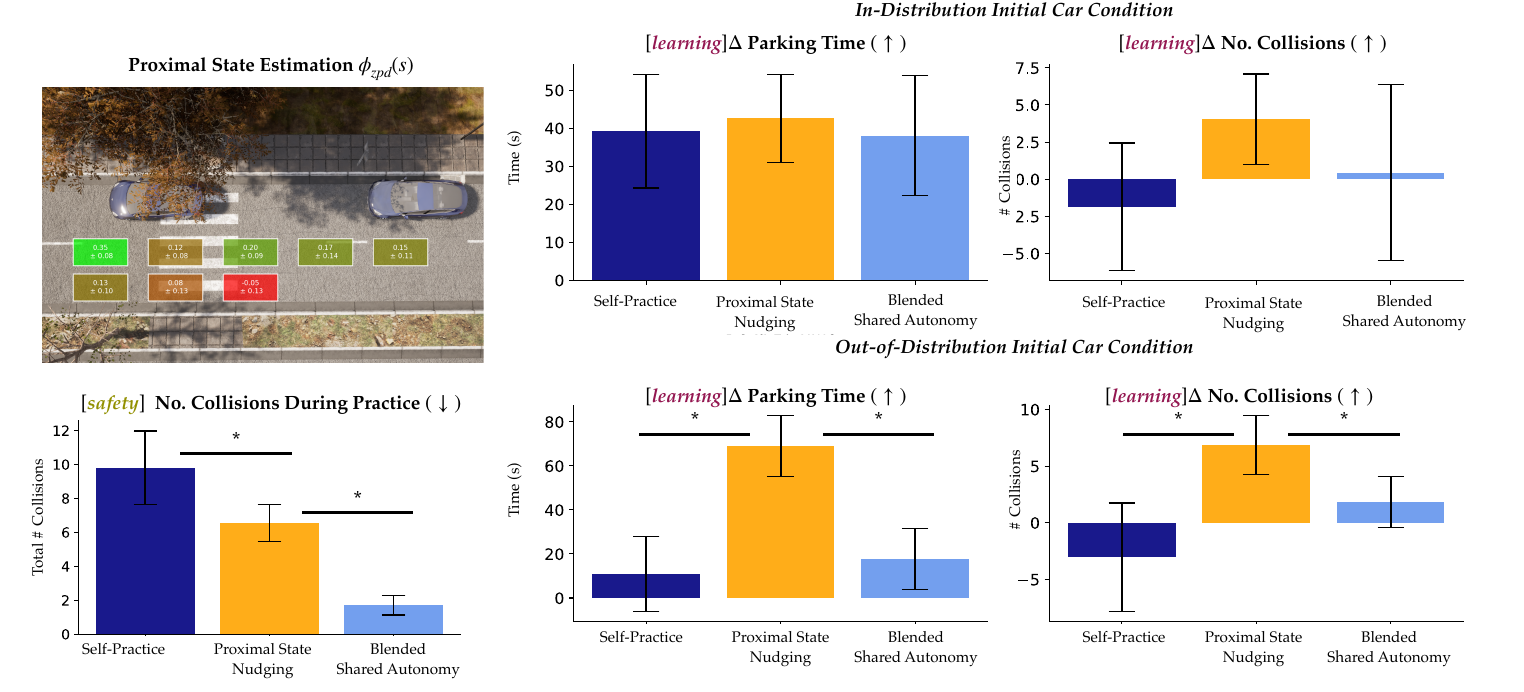}
    \caption{\textbf{Parallel Parking Results} (Top Left) Estimated learnability $\phi_{\mathrm{zpd, t=1}}$ over discretized approach states near the target parking spot; green indicates higher. (Bar Plots)  \psn{}  balances the trade-off between learning gains (e.g. improved time and collisions) and safety (total \# of collisions during practice). in comparison to both blended shared autonomy and unassisted self-practice. Bars show mean standard error,  stat. sig. via two-sample Welch’s t-tests.}
    \label{fig:results-parking}
\end{figure*}

Parallel parking is another challenging driving task requiring precise spatial reasoning and planning, often studied due to its difficulty for human drivers and clear success criteria \cite{chai2023parking}. The task consists of an approach phase followed by a parking phase, and we assume learning difficulties primarily arise from variation in the endpoint of the approach phase (i.e., the start of the parking phase).

\subsubsection{Task Overview}
We again use the CARLA simulator, this time building a parallel parking scenario in the provided Town15 multilane environment. Each  state includes vehicle position $(x,y,z)$, orientation, and velocity, while continuous actions consist of steering, throttle, and brake. We designated a $13$\,m gap between two stationary vehicles as the target spot, and each episode (a single parking attempt) is evaluated with two metrics: parking time and number of collisions. 

\paragraph{Methods} We compare three methods:
\begin{itemize}
\item \textbf{Blended Shared Autonomy (Baseline):} We blend $\pi_{student}$, the user's input controls, and $\pi_{agent}$, a pure-pursuit controller that takes a goal state as input, with $\alpha=0.8$ following Equation \ref{eq:stdsa}.

    \item \textbf{Proximal State Nudging (\psnshort{})} : We sample $B=8$ different states for the start of the parking phase, and we set the state with the highest $\phi_{zpd}$ estimate as the goal for a pure pursuit controller that guides the student. Following Equation \ref{eq:psn}, we set $R$ to a fixed constant, and $w_1=.5$ and $w_2=.5$. 
    \item \textbf{Unassisted Self-Practice (Baseline)}: The student repeatedly practices parking with full control of the car.
\end{itemize}

\subsubsection{Technical Implementation}

\paragraph{Assistive Control Design}
For our assistive control,  $\pi_{expert}$ uses a non-linear Model Predictive Control (NMPC) policy: a proportional heading-and-speed controller for the approach, followed by a finite-horizon nonlinear program (solved with IPOPT) for parking \cite{WachterBiegler2006}. The NMPC controls are blended with human inputs using $\alpha=0.8$, and the agent is queried every 8 frames for tractability.

\paragraph{Proximal State Estimation}
 We discretize the high-level state space near the parking spot and estimate ZPD scores for each state. We collect 272 trajectories from 17 novice drivers, each attempting 4 trials from 8 spots (2 assisted, 2 unassisted, randomized). Following Section~\ref{sec:formalism}, $\phi_{zpd}(s)$ is computed as the average difference of $\rho_{sa}(s)$ and $\rho_{ua}(s)$, which denote average parking time for assisted and unassisted trials . Figure~\ref{fig:results-parking} (left) shows the state directly in front of the target spot has the highest estimated learnability.

\subsubsection{Human Subject Study Results}
We compare  all three methods with the same trial order:
\begin{enumerate}[noitemsep, topsep=0pt]
    \item 2 \textbf{baseline trials} with unassisted user control.
    \item 4 \textbf{practice trials} with either (i) Blended Shared Autonomy, (ii) Unassisted Self-Practice, or (iii) \psn{}.
    \item 2 \textbf{evaluation trials} with unassisted user control (in distribution starting angle)
    \item 2 \textbf{evaluation trials} with unassisted control  (out of distribution starting angle not seen during baseline trials)
\end{enumerate}

In each trial, participants were instructed to park as quickly as possible while avoiding collisions. We recruited 30 students, evenly split between conditions. We compute two learning metrics (parking time and \# of collisions) as the difference between average baseline and evaluation performance, so larger values indicate stronger learning gains. For safety, we measure the total number of collisions during practice trials, with lower values indicate safer learning. 

Our results (see Figure~\ref{fig:results-parking}),  show that \psn{} yields greater improvement than both unassisted self-practice and blended shared autonomy in  improving parking time and reducing collisions, supporting our hypothesis for \textbf{RQ1}. However, statistically strong effects were primarily observed for the randomized-angle out-of-distribution starting angle ($>$ 7x), which  suggests that \psnshort{}'s effect is more salient in conditions that are unfamiliar to a student.  Furthermore, users assigned \psnshort{} had significantly fewer collisions during the practice trials than Unassisted Self-Practice ($<$ 60\%) which validates our hypothesis for \textbf{RQ2}, although Blended Shared Autonomy resulted in an even more significantly safer practice. 

Participants who practices without assistance mainly reported self-learning steering wheel sensitivity, while \psnshort{} participants more often described learning new parking strategies (e.g., \textit{``I came to better understand where I should turn my wheel.''}). Overall, the results for our Parallel Parking task support our prior experimental results in affirming the goal of \psn{} to provide assistance that can ensure safety without compromising users' learning, thereby mitigating skill atrophy.

\section{Discussion and Future Work}
Across simulated agents in \texttt{LunarLander} and $n=60$ human participants in two CARLA driving tasks, \textsc{Proximal State Nudging} resolved the assistance--learning trade-off incurred by prior shared autonomy methods: up to $7\times$ stronger unassisted skill gains than blended shared autonomy, with $50\%$ fewer collisions than self-practice. To our knowledge, these are the first human-subject results for learning-compatible shared autonomy.

Like all shared autonomy methods~\cite{dragan2013policy, reddy2018shared}, PSN assumes access to an expert policy $\pi_{\text{expert}}$; we inherit no stronger requirement than the baselines we compare against. Jointly learning $\pi_{\text{expert}}$ alongside $\phi_{\mathrm{zpd}, t}$, or updating $\phi_{\mathrm{zpd}, t}$ online during deployment (as we do in simulation), are natural next steps, as are longer studies to further test long-term retention.

While we focus on rapid cyber-physical control where skill atrophy is most safety-critical, the underlying formulation---estimating state-conditioned learnability and planning assistance toward it---is domain-agnostic. Potential extensions can include cognitive domains such as AI-assisted coding~\cite{perry2023code}, where assistance granularity (a token, a line, a function) provides a natural mechanism for \psnshort{}.

\bibliography{example_paper,sample}
\bibliographystyle{IEEEtran}


\end{document}